\documentclass[journal,comsoc]{IEEEtran}
\usepackage[T1]{fontenc}% optional T1 font encoding
\usepackage{graphicx}
\usepackage{float}
\usepackage{subfig}
\usepackage{mwe}
\usepackage{mathtools}
\usepackage{multirow}
\usepackage{color, colortbl}
\usepackage{hhline}
\usepackage{makecell}
\usepackage{afterpage}
\makeatletter
\newcommand{\thickhline}{%
    \noalign {\ifnum 0=`}\fi \hrule height 1pt
    \futurelet \reserved@a \@xhline
}
\newcolumntype{"}{@{\hskip\tabcolsep\vrule width 1pt\hskip\tabcolsep}}
\makeatother
\definecolor{Gray}{gray}{0.9}
\usepackage{array}
\usepackage{cite}
\usepackage{url}

\usepackage{amsmath}
\usepackage[cmintegrals]{newtxmath}

\begin{document}
\title{ADSaS: Comprehensive Real-time Anomaly~Detection~System}

\author{Sooyeon Lee,
        and Huy~Kang~Kim\thanks{S. Lee and H. K. Kim are with Korea University.}}

% of Electrical and Computer Engineering, Georgia Institute of Technology, Atlanta,
% GA, 30332 USA e-mail: (see http://www.michaelshell.org/contact.html).}% <-this % stops a space
% \thanks{J. Doe and J. Doe are with Anonymous University.}% <-this % stops a space
% \thanks{Manuscript received April 19, 2005; revised August 26, 2015.}}


\maketitle

\begin{abstract}
Since with massive data growth, the need for autonomous and generic anomaly detection system is increased. However, developing one stand-alone generic anomaly detection system that is accurate and fast is still a challenge. In this paper, we propose conventional time-series analysis approaches, the Seasonal Autoregressive Integrated Moving Average (SARIMA) model and Seasonal Trend decomposition using Loess (STL), to detect complex and various anomalies. Usually, SARIMA and STL are used only for stationary and periodic time-series, but by combining, we show they can detect anomalies with high accuracy for data that is even noisy and non-periodic. We compared the algorithm to Long Short Term Memory (LSTM), a deep-learning-based algorithm used for anomaly detection system. We used a total of seven real-world datasets and four artificial datasets with different time-series properties to verify the performance of the proposed algorithm. 
\end{abstract}

\begin{IEEEkeywords}
anomaly detection, SARIMA, STL, real-time, data stream
\end{IEEEkeywords}

\IEEEpeerreviewmaketitle

\section{Introduction}
\IEEEPARstart{E}{xtremely} vast data leads to severe challenges to a security administrator who should catch all the anomalies in real-time. Anomaly detection cannot be regarded as a human-work anymore. To automate the anomaly detection process, machine-learning-based and statistics-based anomaly detection have been researched within diverse research areas including network intrusion detection, fraud detection, medical diagnoses, sensor events and others. Despite the variety of such studies in recent years, most anomaly detection systems find anomalies in limited conditions. This is because not only a variety of attacks but also multiple sensors in a single device generate a different type of time-series. In the case of IoT devices, which are embedded with multiple sensors, it is inefficient to use independent anomaly detection algorithms to each different sensor. Anomaly detection systems that are resistant to various datasets should detect anomalies autonomously irrespectively of the time-series properties.

Also, the anomaly detection system should occupy as little memory as possible. Most of deep learning based anomaly detection systems are generally unsuitable for an environment such as IoT devices because of the limited memory and light capacity. As most anomalies cause a critical problem to the medical system such as ventricular assist system, anomaly detection system must get less latency. Although there are many types of researches based on deep learning recently, deep learning approaches are not suitable to process data in real-time that changes frequently or has large data since it takes a lot of time to build a model compared to other algorithms.

The critical conditions of anomaly detection system for IoT devices or cloud network intrusion detection are as follows. First is accuracy, second is speed, third is the small size of the model and the last is domain universality~\cite{ref_first,ref_htm}. We propose ADSaS, anomaly detection system using SARIMA and STL, to meet the four conditions. SARIMA is conventional time-series forecast method, and STL is a versatile and robust method for time-series decomposition. Aforementioned methods are commonly used for stationary and periodic time-series data~\cite{ref_tsa}. In our experiments, however, integrating two methods shows better performance not only for periodic data but also for non-periodic data. Moreover, the size of the model and speed are optimized by undersampling and interpolation. For accuracy, we defined an anomaly window for evaluation and then judged how well ADSaS finds anomalies in various datasets.

The contributions of our system are as follows:
\begin{itemize}
\item Regardless of time-series properties, ADSaS detects anomalies with high precision and recall. We verify this by using the time-series from a variety of sources. With the development of Cyber-Physical Systems (CPS) or IoT devices, anomaly detection systems must detect anomaly autonomously and generically for applications.
\item ADSaS detects various types of anomaly. (i.e., peak, dip, concept drift, contextual anomalies, and collective anomalies)
\item ADSaS detects anomalies with short latency. We use two conventional time-series analysis methods and advance performance by undersampling. By undersampling time-series, time-series model is built much faster. Though undersampling causes loss of data, STL recovers that loss by decomposing prediction errors.
\item ADSaS proceeds anomaly detection in real-time for every data stream. 
\end{itemize}

\section{Related works}

In particular, there have been studies such as automotive IDS~\cite{ref_han}, SCADA, control network~\cite{ref_yu} to detect anomalies for mission-critical and safety-critical systems. It is important to develop anomaly detection algorithm robustly for the efficiency of intrusion detection in a modern network environment such as cloud computing~\cite{ref_kwon}. Anomaly detection in time-series is roughly divided to clustering-based approach~\cite{ref_den1, ref_den2} and forecast-based approach. Most of the forecast-based approaches perform anomaly detection based on the error with the predicted value. 

Several machine learning techniques were introduced so far for anomaly detection system. LSTM network has been demonstrated to be particularly useful for anomaly detection in time-series~\cite{ref_lstm}. Jonathan \textit{et al.}~\cite{ref_cpsrnn} also presented a novel anomaly detection system to detect cyber attacks in CPS by using unsupervised learning approach, Recurrent Neural Network (RNN). Sucheta \textit{et al.}~\cite{ref_ecg} applied RNN and LSTM to detect anomalies in ECG signals. They used only a single data source, so did not show the generality of algorithms. 

Some studies used diverse dataset sources to evaluate anomaly detection algorithm. Numenta used Hierarchical Temporal Memory (HTM) algorithm to detect anomaly detection capable for stream time-series~\cite{ref_htm}. HTM is a neural network, and every neuron in HTM remember and predict the value by communicating with each other. Since it is composed of a higher order than other neural networks, it may not be suitable for anomaly detection systems which require high speed. Yahoo suggested EGADS~\cite{ref_egads}, plug-in-out anomaly detection framework, and they indicated that it is essential to use time-series features for anomaly detection. EGADS offers AR, MA, and ARIMA. Several studies~\cite{ref_arima1,ref_arima2} used ARIMA models to forecast time-series, but they did not process errors for a non-periodic dataset. SARIMA was also frequently used for time series prediction, but it was not applied to anomaly detection system~\cite{ref_sarima1, ref_sarima2, ref_sarima3}.

\section{Backgrounds}
\subsection{Time-series analysis}
\subsubsection{Power Spectral Density} Power spectral density is a simple but powerful method to find the frequency of the data~\cite{ref_psd}. Power spectral density of the signal (time-series data) describes the distribution of power which refers to frequency. Power spectral density graph shows clear peaks when the signal has evident frequencies.

\subsubsection{Dickey-Fuller Test} Dickey-Fuller Test tests the null hypothesis that a unit root is present in an autoregressive model~\cite{ref_difu}. The unit root test is carried out under the null hypothesis test value $\gamma=0$ against the alternative hypothesis of $\gamma<0$. The unit root test is an analytical method for determining stationarity of the time-series. In ADSaS, when the p-value of the test is bigger than 0.0005, we reject the hypothesis and refer the time-series as non-stationary data. 

\subsubsection{STL} STL is an algorithm developed to decompose a time-series into three components namely: the trend, seasonality, and residuals (remainder)~\cite{ref_stl}. A trend shows a persistent increasing or decreasing direction in data, seasonality shows seasonal factors over a fixed period, and residuals mean noise of the time-series. For time-series analysis, residuals mainly considered as errors. In this paper, we use residuals of time-series to extract errors that are related to anomalies.

\subsection{Time-series forecast model}
\subsubsection{Autoregressive (AR) Model} AR model is used when a value from a time-series is regressed on previous values from the same time-series. When time-series data has white noise $\alpha_t$, autoregressive parameter $\phi$, an $AR(p)$ model $Z_t$ at time $t$ is defined as:

\begin{equation}
Z_t = \phi_1Z_{t-1}+\phi_2Z_{t-2}+\cdots+\phi_pZ_{t-p}+\alpha_t
\end{equation}

\subsubsection{Moving Average (MA) Model} MA model uses complicated stochastic structure to model time-series~\cite{ref_arma}. When time-series has white noise $\alpha_t$, parameters of the model $\theta$, a $MA(q)$ model $Z_t$ at time $t$ is defined as:

\begin{equation}
Z_t = \alpha_t - \theta_1\alpha_{t-1}-\cdots-\theta_q\alpha_{t-q}
\end{equation}

\subsubsection{ARIMA} ARIMA model generalizes an ARMA model (AR+MA) by replacing the difference among previous values. An ARMA model is applicable only for stationary time-series, ARIMA is applicable for non-stationary time-series. $ARMA(p,q)$ model is given by:

\begin{equation}
Z_t - \phi_1Z_{t-1}-\cdots-\phi_pZ_{t-p} = \alpha_t + \theta_1\alpha_{t-1}+\cdots-\theta_q\alpha_{t-q}
\end{equation}
\begin{equation}
\left(1-\sum_{i=1}^p {\phi_iL^i}\right)Z_t = \left(1+\sum_{i=1}^q {\theta_iL^i}\right)\alpha_t
\end{equation}

In here, $L$ is the lag operator of $Z$. $ARIMA(p,d,q)$ model has parameters $p$ (the order of AR model), $q$ (the order of MA model) and also $d$ (the degree of differencing). When two out of the three parameters are zeros, the model is referred to as AR or MA or I. (i.e., ARIMA(1,0,0) is AR(1)) $ARIMA(p,d,q)$ model is defined as:
\begin{equation}
\left(1-\sum_{i=1}^p {\phi_iL^i}\right)(1-L)^dZ_t = \left(1+\sum_{i=1}^q {\alpha_iL^i}\right)\alpha_t
\end{equation}

\subsubsection{SARIMA} SARIMA is a much more efficient model to express time-series with seasonality than ARIMA model. It has an additional parameter seasonal order called $s$. SARIMA is defined as $SARIMA(p,d,q)(P,D,Q)_s$. The parameters $p, d, q$ are for non-seasonal part of the time-series, and $P, D, Q$ are for seasonal part of the model. In other words, SARIMA creates models with both seasonal and non-seasonal data. For $s=12$, SARIMA builds a time-series model with seasonality per 12 data points. 

\section{Methodology}
\subsubsection{ADSaS}
%Fig.~\ref{fig:fig1} shows how ADSaS determines anomalies. 
ADSaS consists of three modules, dataset analysis module, forecasting module and error processing module. Let the vector $x_t$ is the value of a system at time $t$. Real-time anomaly detection system should classify whether the value is an anomaly or not without using any data after the time ${t}$. First of all, by analyzing the proper size of the train set, dataset analysis module finds the frequency and stationarity of the given dataset. Then, forecasting module forecasts $x_{t+1}, x_{t+2},x_{t+3},$\ldots by using train set (The size of forecast can be changed). When data stream $x_{t+1}$ comes, error processing module calculates the residuals of the error and the cumulative probability of the residuals. If the cumulative probability is less or bigger then the threshold, ADSaS classifies the value as an anomaly and alert. For more accurate forecast model, only the normal value is fed back to the train set. 

\subsubsection{Data analysis module}
STL and SARIMA, mainly used algorithms, work based on the properties of the time-series data. To get the properties, we use \textit{Dickey-Fuller test} for stationarity and \textit{power density spectra} for frequency. When data does not have stationarity, we use one day for default frequency.

\begin{figure}[!t]
\centering
\includegraphics[width=3.0in]{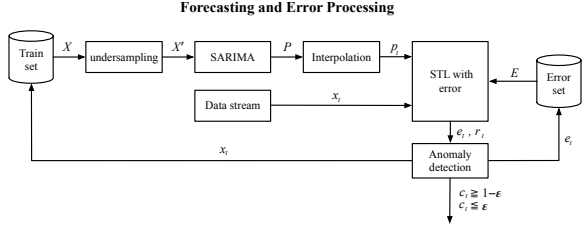}
\caption{The figure shows how the forecasting module and error processing module works.} \label{fig:fig2}
\end{figure}

\subsubsection{Forecasting module}
SARIMA model has a $s$ parameter that represents the seasonality frequency. If a time-series has regular change per one second and repeats every day, $s$ should be at least 86400 to define time-series model. However, large $s$ causes a huge amount of time, which is problematic for practical anomaly detection systems. The ADSaS uses undersampling and interpolation to shorten building time. Fig.~\ref{fig:fig2} is details about how the prediction and error processing works. First, we undersample the train set $X$ to $X' (|X|\gg|X'|)$. If the dataset is recorded at the five-minute interval, we adjust it at the one-hour interval by averaging them. Then, SARIMA model for train set is built to describe and forecast time-series. The interval of the model is one-hour, so we interpolate forecasts at the initial interval (in this case, five-minute) by using \textit{cubic spline interpolation}. Where the predicted value is $p_t$ and real value is $x_t$, absolute prediction error $e_t$ is defined as $p_t - x_t$.

\begin{figure}
\includegraphics[width=3.2in]{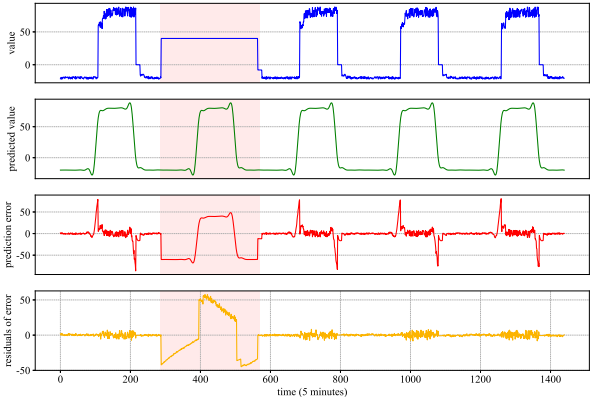}
\caption{Observed value, predicted value, prediction error, and residuals of error for the given time series data.} \label{fig:fig3}
\end{figure}

\subsubsection{Error processing module}
In the error processing module, prediction errors are decomposed by STL and the residuals are calculated. Although undersampling and interpolation arise a serious problem of missing actual data points, regularity of the errors due to lost data points diminishes the residuals. We model the residuals distribution as a rolling normal distribution, though the distribution of prediction errors is not technically a normal distribution. Where the sample mean $\mu$ , and variance $\sigma^2$ are given, the cumulative distribution function is calculated as follows:
\begin{equation}
F(x) = \int_{-\infty}^{x} \frac{e^{-\frac{{(x-\mu)}^2}{2\sigma^2}}}{\sigma\sqrt{2\pi}}\, dx
\end{equation}
We threshold $F(r_t)$ based on a user-defined parameter $\varepsilon$ to alert anomalies\footnote{We used $\varepsilon=0.0005$ for experiments.}. If $F(r_t)$ is smaller than $\varepsilon$ or greater than $1-\varepsilon$, it is determined as anomaly.

Fig.~\ref{fig:fig3} is an example of error residuals\footnote{Anomalies are colored with red.} The error increment is occurred in the normal data (first jump) due to the undersampling and interpolation. However, it is judged to be a regular error by STL, so decomposed to trend or seasonality, no residuals. As the anomaly occurs, the residuals decreases/increases sharply. This causes the dramatic difference in residuals between regular errors and unexpected errors, so anomalies are detected by ADSaS easily.

\section{Experimental Evaluation}
\subsection{Dataset}
There are 11 datasets we used in the experiment, eight datasets from Numenta Anomaly Benchmark (NAB)~\cite{ref_nab} and three datasets from the P corporation, Korea's leading third-party online payment solution. NAB is a benchmark for evaluating anomaly detection algorithms, and it is comprised of over 50 labeled artificial and real-world datasets. Also, the real-world datasets from P are user login statistics, tracks the browser, service provider and login result status. The anomalies are labeled when the real attack attempts are held. All the anomalies were confirmed by P corporation. 

\begin{figure*}[ht]
\centering
\includegraphics[width=6in]{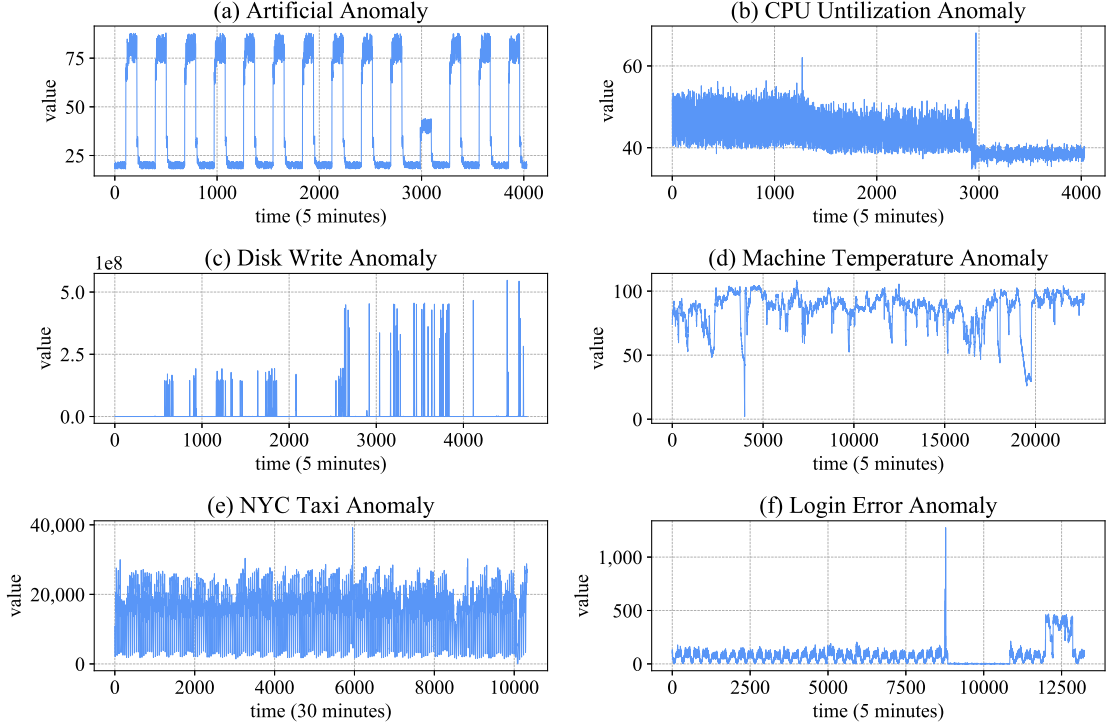}
\caption{The examples of the datasets.} \label{fig:fig4}
\end{figure*}

All datasets except four datasets (NAB artificial jump datasets) are real-world datasets, which cover various fields, including CPU utilization, machine temperature, and user login statistics. The examples of datasets are shown in Fig.~\ref{fig:fig4}. Each dataset has different time series characteristics and anomaly types. For instance, dataset (b) has concept drift anomaly that it should not be detected as anomaly after the drift point. Dataset (c) disk write anomaly has lots of noises. NYC taxi dataset shows various anomalies (peak, dip and partial decrease). 
% The results of dataset analysis using dickey-fuller test and power spectral density is shown in Table \ref{tab1}. 
% By our methodology, only NAB machine temperature dataset has 2 frequency, because it has p-value smaller than 0.0005 and frequency 2. On the other hand, NAB NYC taxi has 40 frequency, but its p-value is larger than 0.0005, so we use default value 7.

\begin{figure}
\centering
\includegraphics[width=3.0in]{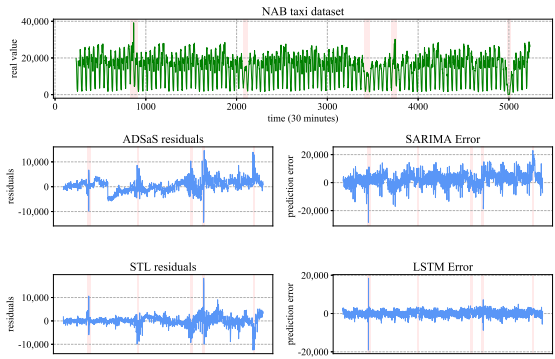}
\caption{Residuals or error of each algorithm for NAB taxi dataset. Anomaly windows are colored with red.} \label{fig:fig5}
\end{figure}

\begin{table*}[ht]
\centering
\caption{Comparison of results between algorithms.}\label{tab2}
\begin{tabular}{l|r|r|r|c|c|c|c|c}
\thickhline
\rowcolor{Gray}
\multicolumn{1}{c|}{{\bfseries Dataset}}
& \begin{tabular}{@{}c@{}}{\bfseries Total} \\ {\bfseries Window}\end{tabular} 
& \begin{tabular}{@{}c@{}}{\bfseries Anomaly} \\ {\bfseries Window}\end{tabular}
& \multicolumn{1}{c|}{{\bfseries Metrics}}
&\begin{tabular}{@{}c@{}}{\bfseries STL} \\ {\bfseries only}\end{tabular}
&\begin{tabular}{@{}c@{}}{\bfseries SARIMA} \\ {\bfseries only}\end{tabular}
&\begin{tabular}{@{}c@{}}{\bfseries LSTM} \\ {\bfseries only}\end{tabular}
&\begin{tabular}{@{}c@{}@{}}{\bfseries LSTM}\\ {\bfseries with}\\ {\bfseries STL}\end{tabular}
&{\bfseries ADSaS}\\
\hline
\multirow{3}{*}{NAB Jumps} & \multirow{3}{*}{335} & \multirow{3}{*}{11-25} & Precision 
& 0.920 & 0.518 & 0.324 & 0.278 & 1.000\\
&&&Recall& 0.910 & 0.727 & 0.500 & 0.750 & 1.000\\
&&&$F_1$-score& 0.903 & 0.583 & 0.370 & 0.392 & {\bfseries 1.000}\\
\hline
\multirow{3}{*}{NAB CPU} & \multirow{3}{*}{335} & \multirow{3}{*}{4} & Precision 
& 0.800 & 0.143 & 0.833 & 0.308 & 1.000\\
&&&Recall& 1.000 & 0.250 & 1.000 & 1.000 & 0.250\\
&&&$F_1$-score& 0.889 & 0.182 & {\bfseries 0.909} & 0.471 & 0.400\\
\hline
\multirow{3}{*}{NAB Disk} 
& \multirow{3}{*}{394} & \multirow{3}{*}{1} & Precision 
& 0.025 & 0.026 & 0.049 & 0.018 & 1.000\\
&&&Recall& 1.000 & 1.000 & 0.222 & 1.000 & 1.000\\
&&&$F_1$-score& 0.049 & 0.051 & 0.049 & 0.018 & {\bfseries 1.000}\\
\hline
\multirow{3}{*}{\begin{tabular}{@{}l@{}}{NAB} \\ {Temperature}\end{tabular}}
& \multirow{3}{*}{315} & \multirow{3}{*}{9} & Precision 
& 0.250 & 0.000 & 0.049 & 0.059 & 1.000\\
&&&Recall& 0.222 & 0.000 & 0.222 & 0.625 & 0.500\\
&&&$F_1$-score& 0.235 & 0.000 & 0.080 & 0.108 & {\bfseries 0.667}\\
\hline
\multirow{3}{*}{NAB Taxi} & \multirow{3}{*}{214} & \multirow{3}{*}{9} & Precision 
& 0.533 & 0.000 & 0.176 & 0.161 & 1.000\\
&&&Recall& 0.889 & 0.000 & 0.333 & 1.000 & 1.000\\
&&&$F_1$-score& 0.667 & 0.000 & 0.231 & 0.277 & {\bfseries 1.000}\\
\hline
\multirow{3}{*}{P Login} & \multirow{3}{*}{1,102} & \multirow{3}{*}{245} & Precision 
& 0.970 & 1.000 & 0.962 & 0.968 & 1.000\\
&&&Recall& 0.922 & 0.307 & 0.307 & 1.000 & 1.000\\
&&&$F_1$-score& 0.945 & 0.470 & 0.466 & 0.984 & {\bfseries 1.000}\\
\hline
\multirow{3}{*}{P Browser} & \multirow{3}{*}{1,102} & \multirow{3}{*}{131} & Precision 
& 0.947 & 1.000 & 1.000 & 0.942 & 1.000\\
&&&Recall& 0.954 & 0.588 & 0.924 & 0.992 & 1.000\\
&&&$F_1$-score& 0.951 & 0.740 & 0.960 & 0.967 & {\bfseries 1.000}\\
\hline
\multirow{3}{*}{P Provider} & \multirow{3}{*}{1,102} & \multirow{3}{*}{141} & Precision 
& 0.914 & 1.000 & 0.917 & 0.849 & 1.000\\
&&&Recall& 0.979 & 0.057 & 0.936 & 1.000 & 0.993\\
&&&$F_1$-score& 0.945 & 0.107 & 0.926 & 0.919 & {\bfseries 0.996}\\
\thickhline
\end{tabular}
\end{table*}

\begin{table}
\centering
\caption{The latency of classification for each stream and model-build.}\label{tab3}
\begin{tabular}{ccc}
\thickhline
\rowcolor{Gray}
\multicolumn{1}{c|}{{\bfseries Algorithm}} 
&  \multicolumn{1}{c|}{{\bfseries Model build(s)}}
&\multicolumn{1}{c}{{\bfseries classify(s)}}  \\
\hline
% Title (centered) &  {\Large\bfseries Hello} & 14 point, bold\\
% 1st-level heading &  {\large\bfseries 1 Introduction} & 12 point, bold\\
% 2nd-level heading & {\bfseries 2.1 Printing Area} & 10 point, bold\\
% 3rd-level heading & {\bfseries Run-in Heading in Bold.} Text follows & 10 point, bold\\
% 4th-level heading & {\itshape Lowest Level Heading.} Text follows & 10 point, italic\\
STL & \multicolumn{1}{|r|}{0.000} & \multicolumn{1}{r}{0.189} \\
SARIMA & \multicolumn{1}{|r|}{3.776} & \multicolumn{1}{r}{0.000}\\
LSTM & \multicolumn{1}{|r|}{1982.410} &\multicolumn{1}{r}{0.001}\\
ADSaS & \multicolumn{1}{|r|}{3.992} & \multicolumn{1}{r}{0.187}\\
\thickhline
\end{tabular}
\end{table}

\subsection{Evaluation Metrics}
We use precision, recall and $F_1$-score to evaluate the algorithm. When the actual anomaly is classified as an anomaly, it is true positive. False positive is when the normal data is classified to be an anomaly. False negative is when the anomaly is classified as normal, and true negative is when the normal data is classified as normal. Depending on the area, false positive may be more important than false negative to check performance or vice versa. We use $F_1$-score to evaluate both precision$(\frac{TP}{TP + FP})$, which is an indicator of whether the anomalies detected by the algorithm is trusty and recall$(\frac{TP}{TP + FN})$, which is an indicator of how many anomalies are detected by the algorithm. The metric of $F_1$-score is \vspace{1mm}$2 \times \frac{Precision \times Recall}{Precision + Recall}$ .

\subsubsection{Anomaly window} We also define \textit{anomaly window} to evaluate the performance of the algorithm. An anomaly may occur only at a certain point (peak, dip), but it may occur over a long period. It is not false positive to detect anomaly at the point immediately before or after the occurrence of an anomaly. It is essential to set the appropriate anomaly windows covering anomalies. Numenta defines their own anomaly windows for NAB dataset, but it is too large to distinguish whether the classification is right or wrong. We use small enough window size to prevent inept detection considered as true positive or true negative. 
In the case of anomalies occurring over a long period, it can be judged as a section composed of several anomaly windows. If the anomaly window is successfully detected in section, it is considered to be true positive after the detection.

\section{Results}

Table~\ref{tab2} shows the comparisons of precision, recall and $F_1$-score for algorithms from different kinds of sources. It shows that ADSaS yields the best overall datasets except one, NAB CPU. For NAB CPU, the algorithm that uses only LSTM shows the best result. LSTM algorithm, however, is a deep learning based algorithm that takes a lot of time to learn. In addition, for data such as NAB Jumps and P Login, which has periodicity and stationary, LSTM shows lowest $F_1$-score than other algorithms. Even LSTM with STL, $F_1$-score is slightly increased in periodic time-series, but it is decreased by about half in case of non-periodic data (NAB CPU).

We note that in most datasets, an algorithm that uses only STL has the next highest $F_1$-score after ADSaS. In particular, as opposed to LSTM, it performed well for NAB Jumps and P Login which are periodic time-series. However, STL does not forecast anything, so it is impossible to automatically correct anomalies to normal values, which is possible in other algorithms. SARIMA only algorithm does not perform well in anomaly detection because its forecast accuracy is compromised by the undersampling and interpolation processes. For NAB disks, which is the noisiest dataset, all but ADSaS has very low $F_1$-score less than 0.05. This suggests that SARIMA, STL, and LSTM cannot handle noise alone, but combining SARIMA and STL shows remarkable performance at handling noise. 

We also analyze the reasons why ADSaS performed poorly on NAB CPU dataset. ADSaS found only one of the four anomaly windows, which is the last anomaly window, the actual concept drift. ADSaS is unable to determine the first anomaly whether or not it is an anomaly because it is used as train set. This is a fatal disadvantage of ADSaS. Both SARIMA and STL require a data set of a certain size to be used as a train set to forecast or decompose time-series. ADSaS uses both algorithms, so the amount of data sets initially used for training is greater than others. However, since there are large enough datasets for anomaly detection in real business, this is not a big problem to ADSaS. In addition, ADSaS shows near-perfect accuracy for most datasets.

Fig.~\ref{fig:fig5} shows examples of the residuals and errors from each algorithm in NAB taxi dataset. The anomaly is determined by the cumulative distribution function of these data. There are five anomaly sections (including peak, dip, partial decrease), and a total number of anomaly windows is nine. First, ADSaS and STL have some similar forms but STL shows some bumps between the fourth and fifth anomaly sections which cause false positives. SARIMA generates a lot of errors regularly due to its uncertainty of forecasting. LSTM shows the lowest prediction error compared to other algorithms, but the prediction error is increased only at the point where peaks exist. For dip and partial decrease, where the value is suddenly reduced, the prediction error is low because LSTM quickly adjusts to the value.

Table~\ref{tab3} is a comparison of latency between algorithms. STL does not need to build a model, so only time-decomposition process increases latency. Unlike other algorithms, the time decomposition of STL is directly linked to anomaly detection and cannot proceed with batch. For SARIMA, the forecast size can be adjusted to help speed up the forecasting. In this experiment, SARIMA model predicts the daily data in advance. Therefore, anomaly classification using SARIMA is very fast because all it has to do is calculate the actual stream data difference. Although we used the LSTM model with 12 neurons, two hidden layer and \textit{relu} activation function in this experiment, which is comparatively not a heavy model, LSTM took the 1982 seconds to build the model. As the number of neurons and hidden layers increases, building or updating LSTM's model takes an extraordinary amount of time.

\section{Conclusions}
By combining STL and SARIMA, we have presented algorithms to detect various anomalies in datasets from various sources. In addition, comparing with LSTM shows that conventional time-series analysis has better performance and accuracy than the deep-learning algorithm. In this paper, we have discussed the forecasting model SARIMA does not give accurate predictions, but STL is able to resolve incomplete predictions by decomposing the prediction errors. It supports the fact that the STL algorithm will be more useful in anomaly detection than other approaches in error processing, such as likelihood. We also showed that the conventional time-series techniques are applicable to noisy and non-stationary datasets (NAB Disk). We applied our algorithm to real online payment system data and showed that ADSaS can be applied directly to the real industry. ADSaS succeeded in detecting anomaly right at the time of the attack. In this experiment, only the SARIMA model is used as the time-series prediction algorithm, but other time-series models including GARCH model, that expresses white noises, can be used as a predictor module. Furthermore, we need to update our algorithms to detect anomaly using multivariate datasets because we conducted the experiments on datasets with an only single variable.

\section*{Acknowledgements}
This work was supported under the framework of international cooperation program managed by National Research Foundation of Korea (No.2017K1A3A1A17 092614).

\end{document}